\pdfoutput=1

\documentclass[11pt]{article}

\usepackage[]{acl}

\usepackage{times}
\usepackage{latexsym}
\usepackage{booktabs}
\usepackage{graphicx}
\usepackage{scalerel}
\usepackage{multirow}
\usepackage{xurl}
\usepackage{tabularx}

\newcolumntype{S}{>{\hsize=.45\hsize}X}
\newcolumntype{s}{>{\hsize=.3\hsize}X}
\newcolumntype{a}{>{\hsize=.2\hsize}X}

\usepackage{pifont}
\newcommand{\cmark}{\ding{51}}%
\newcommand{\xmark}{\ding{55}}%
\usepackage{comment}

\usepackage[T1]{fontenc}

\usepackage[utf8]{inputenc}

\usepackage{microtype}
\usepackage[compact]{titlesec}

\definecolor{Gray}{gray}{0.95}
\usepackage{colortbl}

\title{Neopronouns: to model or not to model?}
\title{Pronouns -- to Model or Not to Model?}
\title{Welcome to the Modern World of Pronouns -- Is it Fair to Model them?}
\title{Identity-Inclusive NLP beyond Gender \\-- Welcome to the Modern World of Pronouns}
\title{Welcome to the Modern World of Pronouns:\\Identity-Inclusive Natural Language Processing beyond Gender}

\author{Anne Lauscher \\
  MilaNLP \\ 
  Università Luigi Bocconi \\ 
  Milan, Italy \\
  \small{\texttt{anne.lauscher@unibocconi.it}} \\ \And
  Archie Crowley \\
  Linguistics \\
  University of South Carolina \\
  Columbia, SC, USA \\
  \small{\texttt{acrowley@sc.edu}} \\
   \And
  Dirk Hovy \\
  MilaNLP \\ 
  Università Luigi Bocconi \\ 
  Milan, Italy \\
  \small{\texttt{dirk.hovy@unibocconi.it}} \\}

\begin{document}
\maketitle
\begin{abstract}
\emph{Trigger warning: This paper contains some examples which might be offensive to some users.}

\vspace{0.5em}
The worls of pronouns is changing. From a closed class of words with few members to a much more open set of terms to reflect identities. However, Natural Language Processing  (NLP) is barely reflecting this linguistic shift, even though recent work outlined the harms of gender-exclusive language technology. Particularly problematic is the current modeling 3rd person pronouns, as it largely ignores various phenomena like neopronouns, i.e., pronoun sets that are novel and not (yet) widely established. This omission contributes to the discrimination of marginalized and underrepresented groups, e.g., non-binary individuals. However, other identity-expression phenomena beyond gender are also ignored by current NLP technology. 
In this paper, we provide an overview of 3rd person pronoun issues for NLP. Based on our observations and ethical considerations, we define a series of desiderata for modeling pronouns in language technology. We  evaluate existing and novel modeling approaches w.r.t.\ these desiderata qualitatively, and quantify the impact of a more discrimination-free approach on established benchmark data.
\end{abstract}

\section{Introduction}
Pronouns are an essential component of many languages and often one of the most frequent word classes. Accordingly, NLP has long studied tasks related to them, e.g., pronoun resolution \citep[e.g.,][]{hobbs1978resolving}. 
Simplistically, they can be defined as \emph{``a word (such as I, he, she, you, it, we, or they) that is used instead of a noun or noun phrase''}.\footnote{Essential definition provided by the Mirriam Webster Online Dictionaly at \url{https://www.merriam-webster.com/dictionary/pronoun}} 
Linguistic studies have pointed out the complexity of pronouns, though~\citep[e.g.,][]{postal1969so, mckay1993term}.
Pronouns can carry demographic information -- in English, for example, information about the \emph{number} of referees and a single referee's (grammatical) \emph{gender}.\footnote{Grammatical, biological, and self-identified gender should not be confounded, but are often treated interchangeably by lay audiences.} 
Even more information can be conveyed by pronouns in other, non-pro-drop languages. Consider  Arabana-Wangkangurru, a language spoken in Australia, in which a speaker uses different pronouns depending on whether the referee is part of the same social or ritual group (moiety)~\citep{hercus1994grammar}. 
As such, pronouns shape how we perceive individuals and can even reflect cultural aspects~\citep[e.g.,][]{doi:10.1177/0022022198293005} and ideologies~\citep[e.g.,][]{muqit2012ideology}. Consequently, pronoun usage should be considered a sensitive aspect of natural language use.

Accordingly, in many western societies, these phenomena have been drawing more and more attention. For instance, in 2020, the American Dialect Society voted \emph{``(My) Pronouns''} as the 2019 Word of the Year and \emph{Singular ``They''} as the Word of the Decade~\citep{wordofyear}. %
Recently, there has been a shift in pronoun usage~\citep{krauthamer2020great}, partially due to shifts in the perception of \emph{gender}, driven by the queer-feminist discourse~\citep[e.g.,][]{butler2002gender,butler2004undoing}. Related to this is the open discussion of identity beyond binary gender. For instance, a person who does not identify their gender within the gender binary (e.g., a \emph{nonbinary} or \emph{genderqueer} person) might use \emph{singular ``they''} as their pronoun. %
Recently, the French dictionary ``Le Robert'' added the non-binary pronoun \emph{``iel''} to its list of words.\footnote{\url{https://dictionnaire.lerobert.com/dis-moi-robert/raconte-moi-robert/mot-jour/pourquoi-le-robert-a-t-il-integre-le-mot-iel-dans-son-dictionnaire-en-ligne.html}}

This ``social push'' to respect diverse gender identities also affects aspects of NLP. Recent studies have pointed out the potential harms from the current lack of non-binary representation in  NLP data sets, embeddings, and tasks~\citep{cao-daume-iii-2021-toward, dev-etal-2021-harms}, and the related issue of unfair stereotyping of queer individuals~\citep{barikeri-etal-2021-redditbias}. 
However, the research landscape on modern pronoun usage is still surprisingly scarce, hindering progress for a fair and inclusive NLP. 

Further linguistic research has identified identity aspects of pronouns beyond gender~\citep{miltersen2016nounself}. Specifically, \emph{nounself} pronouns, functionally turning pronouns from a \emph{closed} to an \emph{open} word class. To the best of our knowledge, these aspects have been completely ignored by NLP so far. We did not find a single work systematically describing all of the currently existing phenomena even just in English third-person pronoun usage (let alone other languages).\footnote{For instance, while we found hits for the Google Scholar query \emph{``neopronoun''}, we did not get any results for variants of \emph{``nameself pronoun''}, %
or \emph{''emojiself pronoun''}.} 
In contrast, a fair number of discussions are taking place in queer Wikis and forums. While it is still unclear which of these phenomena will persist over the next decades, people are using and discussing them, and accordingly, we as a research community should adapt.

\paragraph{Contributions.} In this ``living draft'', \textbf{1)} we are the first to provide a systematic overview of existing phenomena in English 3rd person pronoun usage. Our results will inform future NLP research on ethical NLP and  non-binary representation. We provide the first NLP work acknowledging \emph{otherkin} identities. We support our observations with a corpus analysis on Reddit. \textbf{2)}~Based on our overview, we derive five desiderata for modeling third-person pronouns. Based on these, \textbf{3)} we discuss various existing and novel paradigms for \emph{when} and \emph{how} to model pronouns. \textbf{4)}~Finally, we quantify the impact of discrimination-free non-modeling of pronouns on a widely established benchmark.

\section{Related Work}
While there are some works in NLP on gender-inclusion~\citep[e.g.,][]{dev-etal-2021-harms} and gender bias in static~\citep[e.g.,][\emph{inter alia}]{Bolukbasi,gonen-goldberg-2019-lipstick, DEBIE} and contextualized~\citep[e.g.,][\emph{inter alia}]{kurita-etal-2019-measuring, bordia-bowman-2019-identifying, lauscher2021sustainable} language representations as well as works focusing on specific gender bias in downstream tasks, e.g., natural language inference~\citep{dev2020measuring} and co-reference resolution~\citep[e.g.,][]{rudinger-etal-2018-gender, webster-etal-2018-mind}, we are not aware of any work that deals with the broader field of identity-inclusion. Thus, there is no other NLP work that deals with a larger variety of pronouns and acknowledges pronouns as an open word class. For surveys on the general topic of unfair bias in NLP we refer to \citet{blodgett-etal-2020-language} and \citet{shah-etal-2020-predictive}.
Recently, \citet{dev-etal-2021-harms} pointed broadly at the harms~\citep{barocas2017problem} arising from gender-exclusivity in NLP. They surveyed queer individuals and assessed non-binary representations in existing data set and language representations. In contrast to them, we specifically look at third-person pronoun usage and how to model such phenomena.
\citet{webster-etal-2018-mind} provide a balanced co-reference resolution corpus with a focus on the fair distribution of pronouns but only focus on the gendered binary case. Closest to us, \citet{cao-daume-iii-2021-toward} discuss gender inclusion throughout the machine learning pipeline beyond the binary gender conception. While they are also the first to consider non-binary pronouns, including some neopronouns, in co-reference resolution, they do not acknowledge the broader spectrum of identity-related pronoun phenomena.

\section{A Note on Identity and Pronouns}

This work focuses on the relationship between identity and pronouns. \emph{Identity} refers to an individual's self-concept, relating to the question of what makes each of us unique~\citep{maalouf2011identity}. It can be seen as a two-way process between an individual and others~\citep{donath2002identity}, and relates to different dimensions, e.g., one's gender.

\paragraph{Gender Identity.} 
Gender identity, as opposed to gender expression or sex, is one's subjective sense of gender~\citep{stryker2017transgender}. In this work, we conceptualize gender identities beyond the binary notion (\emph{man}, \emph{woman}), e.g., non-binary gender, transgender, agender, polygender, etc.

\paragraph{Otherkin Identity.} 
Individuals with otherkin identity do not entirely identify as human~\citep{laycock2012we}, e.g., vamp. \citet{miltersen2016nounself} note that otherkin individuals often identify with \emph{nounself} pronouns matching their kin.

\newcite{stryker2017transgender} highlights the strong relationship between gender identity and pronouns. As \newcite{raymond2016linguistic} notes, pronoun choices construct the individual's identity in conversations and the relationship between interlocutors. According to \newcite{cao-daume-iii-2021-toward}, pronouns are a way of expressing referential gender. Referring to an individual with sets of pronouns they do not identify with, e.g., resulting in misgendering, is considered harmful~\citep{dev-etal-2021-harms}.

\section{Phenomena in Third-person Pronoun-Usage}
\setlength{\tabcolsep}{5pt}
\begin{table}[t]\centering
\small
\begin{tabular}{llllll}\toprule
\multirow{2}{*}{\textbf{Nom.}} &\multirow{2}{*}{\textbf{Acc.}} &\textbf{Poss.} &\textbf{Poss.} & \multirow{2}{*}{\textbf{Reflexive}} \\
 & &\textbf{(dep.)} &\textbf{(indep.)} & \\\midrule
\multicolumn{5}{l}{\emph{Gendered Pronouns}} \\
he &him &his &his &himself \\
she &her &her &hers &herself \\

\multicolumn{5}{l}{\rule{0pt}{2.5ex}\emph{Gender-Neutral Pronouns}} \\
they & them & their & theirs & themself \\
\multicolumn{5}{l}{\rule{0pt}{2.5ex}\emph{Neopronouns}} \\
thon &thon &thons &thons &thonself \\
e &em &es &ems &emself \\
ae &aer &aer &aers &aerself \\
co &co &cos &cos &coself \\
ve/ vi &ver/ vir &vis &vers/ virs &verself/ virself \\
xe &xem &xyr &xyrs &xemself \\
ey &em &eir &eirs &emself \\
e &em &eir &eirs &emself \\
ze &zir &zir &zirs &zirself \\
...\\
\multicolumn{5}{l}{\rule{0pt}{2.5ex}\emph{Nounself Pronouns}} \\
star & star & stars & stars & starself\\
vam & vamp & vamps & vamps & vampself\\
...\\
\multicolumn{5}{l}{\rule{0pt}{2.5ex}\emph{Emojiself Pronouns}} \\
\scalerel*{\includegraphics{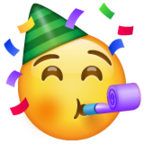}}{\strut} & \scalerel*{\includegraphics{img/partying-face_1f973.png}}{\strut} & \scalerel*{\includegraphics{img/partying-face_1f973.png}}{\strut}s & \scalerel*{\includegraphics{img/partying-face_1f973.png}}{\strut}s & \scalerel*{\includegraphics{img/partying-face_1f973.png}}{\strut}self\\
\scalerel*{\includegraphics{img/partying-face_1f973.png}}{\strut} & \scalerel*{\includegraphics{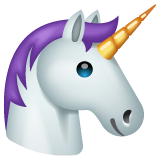}}{\strut} & \scalerel*{\includegraphics{img/unicorn_1f984.png}}{\strut}s & \scalerel*{\includegraphics{img/unicorn_1f984.png}}{\strut}s & \scalerel*{\includegraphics{img/unicorn_1f984.png}}{\strut}self\\
...\\
\multicolumn{5}{l}{\rule{0pt}{2.5ex}\emph{Numberself Pronouns}} \\
$0$ & $0$ & $0$s &$0$s &$0$self \\
$\frac{1}{3}$ & $\frac{1}{3}$ & $\frac{1}{3}$s & $\frac{1}{3}$s & $\frac{1}{3}$self \\ 
...\\
\multicolumn{5}{l}{\rule{0pt}{2.5ex}\emph{Nameself Pronouns}} \\
John & John & Johns & Johns & Johnself \\
... \\
\bottomrule
\end{tabular}
\caption{Non-exhaustive overview of phenomena related to third-person pronoun usage in English.}\label{tab:phenomena}
\end{table}
We describe existing phenomena and analyze their presence in a collection of threads from Reddit.\footnote{\url{https://www.reddit.com}}
\subsection{Existing Phenomena}
 Overall, individuals can choose $n$ sets of pronouns with $n \ge 0$. If $n=0$, the individual does not identify with any singular 3rd person pronoun. If $n>1$, the individual identifies with more than one set of pronouns, possibly each set reflecting overlapping or non-overlapping aspects of their identity. We provide examples of these sets in Table~\ref{tab:phenomena}. Note that this list is non-exhaustive and that the described phenomena are non-exclusive.

\paragraph{Gendered Pronouns.} In English, two sets of standard gendered pronouns are available, \emph{he/him/himself} and \emph{she/her/herself}. %

\paragraph{Gender-Neutral Pronouns.} Given the history of generic singular \emph{they} in English (e.g., Who was at the door? \emph{They} left a note.), there has been an uptake of singular \emph{they} by non-binary individuals as a gender-netural pronoun option\footnote{\url{https://gendercensus.com/results/2021-worldwide-summary/}} \citep{conrod2019they,Konnelly-Cowper-singularthey}. Further, there has been increasing institutional recognition with dictionaries and style guides supporting its use. 

\paragraph{Neopronouns.} As an alternative to the singular \emph{they}, individuals started creating and sharing novel sets of 3rd person pronouns~\citep{mcgaughey2020understanding}. More traditional and rather well-known sets of neopronouns include, e.g., the so-called Spivak pronouns \emph{e/emself}~(used in \citep{spivak1990joy}) and related variations. During our research, we were able to observe various subcategories of neopronouns, partially described in the academic literature.  %

\vspace{0.5em}
\noindent\emph{Nounself Pronouns.} According to \citet{miltersen2016nounself}, nounself pronouns are pronouns that are ``[...] prototypically  transparently  derived  from  a  specific  word, usually  a noun''. Individuals may identify with certain nouns, possibly corresponding to distinct aspects of their identity, e.g., \emph{kitten/kittenself}, \emph{vamp/vampself}. The author notes the difficulty of clearly defining nounself pronouns, neopronouns, and other phenomena. The phenomenon is assumed to have first appeared in 2013.

\vspace{0.5em}
\noindent\emph{Emojiself Pronouns.} Similar to nounself pronouns, individuals may identify with sets of emojis, possibly reflecting different aspects of their identity, e.g., \scalerel*{\includegraphics{img/partying-face_1f973.png}}{\strut}\emph{/} \scalerel*{\includegraphics{img/partying-face_1f973.png}}{\strut}\emph{self}. Emojiself pronouns are intended for written communication. Note, that, at the time of writing this manuscript, there seem to exist no academic description of emojiself pronouns. However, we were able to find evidence of their existence on several social media platforms and wikis, e.g., Tumblr,\footnote{E.g., \url{https://pronoun-archive.tumblr.com/post/188520170831}} MOGAI Wiki,\footnote{\url{https://mogai.miraheze.org/wiki/Emojiself}; according to the article, the origin of emojiself pronouns is unclear but might date back to 2017}  Twitter,\footnote{Example of a user complaining about LinkedIn not allowing for emojiself pronouns in the pronoun field: \url{https://twitter.com/frozenpandaman/status/1412314202119700480/photo/1}} and Reddit.\footnote{E.g., \url{https://www.reddit.com/r/QueerVexillology/comments/p09nek/i_made_a_flag_for_the_emojiself_pronoun_set/}} 

\vspace{0.5em}
\noindent\emph{Numberself Pronouns.} Another form of neopronouns/ nounself pronouns are numberself pronouns. Analogous to before, we assume that here,  the individual identifies or partially identified with a number, e.g., \emph{0/0/0s/0s/0self}.\footnote{\url{https://pronoun-provider.tumblr.com/post/148452374817/i-think-numbers-as-pronouns-would-be-pretty-cool}}

\vspace{0.5em}
\noindent\emph{Nameself Pronouns.} Individuals may identify with pronouns build from their name, e.g., \emph{John}/\emph{Johnself}, overlapping with nullpronomials.\footnote{\url{https://pronoun.fandom.com/wiki/Nullpronominal}} 

\paragraph{Alternating Pronouns.} Given that people can identify with $n>1$ sets of pronouns, the pronouns they identify with can be either equally identified-with sets, or change potentially depending on the context (\emph{mutopronominal}). For instance, individuals who are also performer may use \emph{stage pronouns}. Similarly, genderfluid individual may identify with a certain pronoun at a certain point in time (\emph{pronoun fluidity}, \citep{cherry2020music}). Some individuals identify with the pronouns of the person who is referring to them (\emph{mirrored pronouns}). Other individuals use set(s) of \emph{auxiliary pronouns}, e.g., for situations, in which individuals referring to them have problems with using the most identified-with sets of pronouns (e.g., in the case of emojiself pronouns and oral communication). Note that alternating pronoun sets may be even used in the same sentence for referring to the same individual.\footnote{\url{https://www.reddit.com/r/NonBinary/comments/jasv5r/alternating_pronouns_in_same_sentence/}}   

\paragraph{No Pronouns.} Some individuals do not identify with any pronouns. In this case, some individuals identify most with their name being used to refer to them, nameself pronouns, or avoid pronouns.

\vspace{0.5em}
\begin{figure}[t]
\centering
\includegraphics[width=\linewidth]{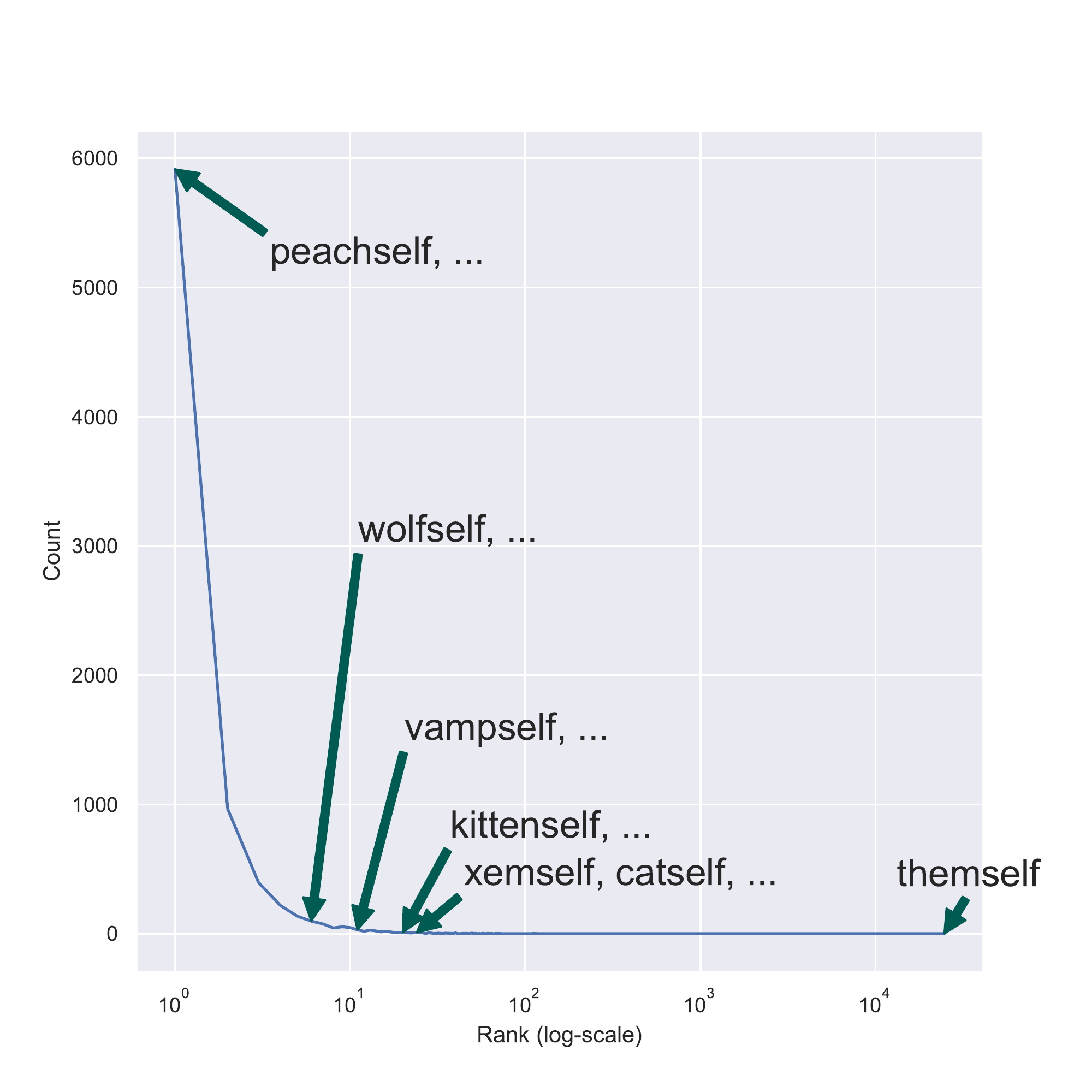}
\caption{Token ranks (log-scale) and rank counts of the tokens returned against our reflexive regular expression pattern from Reddit with example annotations.}
\label{fig:ranks}
\end{figure}

\subsection{Corpus Analysis: Neopronouns in Reddit}
\setlength{\tabcolsep}{4pt}
\begin{table*}[t]\centering
\small
\begin{tabularx}{\linewidth}{a s S X}\toprule
\textbf{Match} &  \textbf{Subreddit} &  \textbf{Thread Title} & \textbf{Thread Excerpt} \\
\midrule
\rule{0pt}{0ex} meowself & monsterhunterrage &  \emph{Fureedom Mewnite can die in my litterbox.} & \emph{I don't like this game. But I still want meowself to play it, meow. Cause it's fun, even though I hate it.} \\

\rule{0pt}{2.5ex}& offmychest & \emph{Neopronouns are going too far.} &  \emph{I get some pronouns like ze/zir, xe/xem, etc. I agree with those. But why are people using ghost/ghostself and meow/meowself? That’s really utter bullshit.} \\
\rule{0pt}{2.5ex} bunself & TiADiscussion & \emph{I am genderfluid, pansexual, and mentally ill. I have a lot of SJW friends. AMA!}& \emph{They/them pronouns are coolest with me, but I won't be angry if you use he or she. You can use bun/buns/bunself, if you are feeling special. (That was a joke.)}  \\
\rule{0pt}{2.5ex} & rpdrcirclejerk & \emph{Xi am so proud to announce that the new word of the year is.....} & \emph{--Cinnagender-- which means you identify with our beloved and innocent cinnamon buns. The pronoun set is cinne/cinns/cinnself or alternatively bun/buns/bunself i am so happy to be a member of a community that ignores the oppressive gender binary, which is a social construct, i.e., it is not real}\\
\rule{0pt}{2.5ex} zirself & mypartneristrans & \emph{Ran into our first roadblock} &  \emph{I asked what I could do to help zir lowering the feeling of disphoria, and ze said ze\'d maybe feel better about zirself if ze\'d drink a tea.} \\
\rule{0pt}{2.5ex} & Negareddit & \emph{No, Redditors. If you're a horrible person online, you're probably a horrible person offline too.} & \emph{Hello folks. Omg. I think this person is about to kill zirself! (emphasis on "zirself". COMEDIC GENIUS)}\\
\bottomrule
\end{tabularx}
\caption{Example neopronouns and corresponding excerpts from Reddit retrieved via our heuristic method. We slightly modified the excerpts to lower searchability and increase the privacy of the users.}\label{tab:examples}
\end{table*}
\paragraph{Setup.} 
We conduct an additional quantitative analysis for the presence of neopronouns in Reddit. To this end, we use Reddit threads created between 2010 and 2021, cleaned by previous work and provided through Huggingface Datasets (127,445,911 lines).\footnote{\url{https://huggingface.co/datasets/sentence-transformers/reddit-title-body}} As we are interested in capturing novel pronouns and as the list of possible pronouns is indefinite, we proxy neopronouns via the suffixes \emph{self} and \emph{selves} indicating the reflexive case and match them through a regular expression. Additionally, we filter out non-3rd person pronouns (e.g., \emph{yourself}, \emph{ourselves}, plural \emph{themselves}) as well as common variations of these (e.g., \emph{urself}) and other common non-pronoun expressions we found in the data (e.g., \emph{do-it-yourself}). This process leaves us with a total of 9,075 unique tokens with in total 74,768 textual mentions. 

\paragraph{Results.} 
An initial manual analysis reveals that, unsurprisingly, many of the matches are false positives, i.e., not real neopronouns (e.g., \emph{non-self}, a common concept in Buddhist philosophy, and \emph{tomyself}, a common spelling mistake of \emph{to myself}). However, our method still finds relevant cases. Examples are depicted in Table~\ref{tab:examples}. Many discussions in which we detect nounself pronouns center on the phenomena themselves, including, e.g., individuals stating that they are interested in using a specific pronoun or individuals stating that they refuse to acknowledge the phenomenon. Some discussions involve people reporting on personal experiences and problems and seeking advice. To obtain a quantitative view of the results, we plot the ranks (i.e., number of occurrences of a token) against their number of tokens (Figure~\ref{fig:ranks}). The result is a highly skewed Zipf's distribution: while the highest ranks appear only once (e.g., \emph{themself} with 24,697 mentions), some tokens appear only a couple of times (e.g., the neopronoun \emph{xemself} with 24 mentions), and the vast majority appears only once (e.g., many nounself pronouns such as \emph{peachself}).

\section{How Can and Should We Model Pronouns?}
We devise five desiderata based on our previous observations, personal experiences, and expert knowledge from interactions with LGBTQIA+ associates. Additionally, we collect informal feedback from individuals who use gender-neutral pronouns. We then assess how well classic and novel pronoun modeling paradigms fulfil the five criteria.

\subsection{Desiderata}
\paragraph{D1. Refrain from assuming an individual's identity and pronouns.} A model should not assume an individual's identity, e.g., gender, or pronouns based on, e.g., statistical cues about an individual's name, also not in a binary gender setup. Only because the name \emph{John} typically appears together with the pronoun \emph{he}, the model should never assume that a person with the name \emph{John} identifies as a man and that every \emph{John} uses the pronoun \emph{he}.

\paragraph{D2. Allow for the existing sets of pronouns as well as for neopronouns.} A model should be able to handle not only the existing set of  "standard" pronouns in a language but also other existing pronouns, e.g., neopronouns.

\paragraph{D3. Allow for novel pronouns at any point in time.} 
On top of \textbf{D2}, a model should allow for novel, i.e., unseen, pronouns to appear at any point in time. This condition is necessary to handle the fact that neopronouns are not a fixed set, but evolving, and because related phenomena (emojiself and nameself pronouns) turn pronouns from a \emph{closed} to an \emph{open class} part of speech. 

\paragraph{D4. Allow for multiple, alternating, and changing pronouns.} 
A model should not assume that the pronoun set for an individuum at time $t$ will be the same as at time $t-1$. Even within the same sequence, pronoun sets might change.

\paragraph{D5. Provide an option to set up individuals' sets of pronouns.} While most NLP models are trained offline and do not interact with the user, some are designed to interact with individuals, e.g., dialog systems. In this context, setting up individuals' sets of pronouns can help avoid harmful interactions (depending on the concrete sociotechnical deployment scenario).

\subsection{Modeling Paradigms}
\label{sec:modeling}
\setlength{\tabcolsep}{8.5pt}
\begin{table}[t]\centering
\small
\begin{tabular}{l l l l l l}\toprule
\textbf{Paradigm} & \textbf{D1} & \textbf{D2} & \textbf{D3} &\textbf{D4} & \textbf{D5} \\
\midrule
Classic & \xmark & \xmark & \xmark & \xmark & \xmark \\
Bucketing & \xmark & \cmark & \cmark & ? & \xmark \\
Delexicalization & \cmark & \cmark & \cmark &  \cmark & \xmark \\
Post-hoc & \cmark & \cmark & \cmark & \cmark & \cmark \\
\bottomrule
\end{tabular}
\caption{Modeling paradigms and how they allow for fullfilling the desiderata D1--D5.}\label{tab:des}
\end{table}
We compare four general modeling paradigms with D1--D5 in Table~\ref{tab:des}.

\paragraph{Classic Statistical Modeling.} Traditionally, pronouns have been treated as a \emph{closed word class}.
Generally, statistical models do not make assumptions about this (except if the vocabulary is manually curated). However, in models exploiting cooccurences, e.g., via word embeddings (GloVe~\citep{pennington2014glove}) or deep language models (BERT~\citep{devlin-etal-2019-bert}, RoBERTa~\citep{liu2019roberta}), the models will likely misrepresent underrepresented pronoun-related phenomena. \newcite{dev-etal-2021-harms} provided an initial insight by showing that singular \emph{they} and the neopronouns \emph{xe} and \emph{ze} do not have meaningful vectors in GloVe and BERT. %

\paragraph{Bucketing.} One option, previously discussed by \newcite{dev-etal-2021-harms}, is to apply bucketing, i.e., to decide on a fixed number of majority classes, e.g., \emph{male} pronouns, \emph{female} pronouns, and one or multiple classes for the ``rest of the pronouns'', e.g., \emph{other}. The advantage of this approach is that it can map existing and novel pronouns to the \emph{other} class. However, it still makes identity assumptions -- and due to unequal representations of \emph{main} and \emph{other} classes, it will inevitably lead to discrimination.

\paragraph{No Modeling -- Delexicalization.} 
Given that the classic approach and bucketing both lead to unfair treatment of underrepresented groups, the alternative is to explicitly not model pronouns in their surface forms. This process, commonly named delexicalization, has proved helpful for other tasks where models capture misleading lexical information, e.g., fact verification~\citep[e.g.,][]{suntwal-etal-2019-importance}, or resource-lean scenarios, e.g., cross-lingual parsing~\citep[e.g.,][]{mcdonald-etal-2011-multi}.\footnote{In fact, accounting for novel pronouns and novel ways of using pronouns is a  resource-lean scenario.} 
In this case, the model is forced to not rely on spurious lexical cues related to gender, e.g., that \emph{John} occurs most often with the pronoun \emph{he}. Instead, the model learns a single representation for all pronouns and relies on other task-related conceptual and commonsense information for disambiguation.

\paragraph{Post-hoc Injection of Modeling Information/ Modeling at Test Time.} 
For human-to-human interactions, several LGBTQIA+ guides recommend to (1) first try generic pronouns (e.g., singular \emph{they}), and (2) switch to other sets of pronouns once the conversation partner communicates them. For uncommon or novel pronouns,  several web pages have explicitly been set up for practising how to use them.\footnote{E.g., \url{https://www.practicewithpronouns.com/\#/?_k=66emp7}} 
In this work, \emph{we propose that NLP systems should work similarly} -- if technically possible and depending on the concrete sociotechnical deployment scenario. To this end, we can use intermediate training procedures~\citep[e.g.,][]{hung2021ds} for pronoun-related model refinement. E.g., we can use synthetic data created through similar procedures as the ones employed on these websites. Another option is only model pronouns at test time, e.g., through simple replacement procedures.

\section{How Much Would We Loose?}
In \S\ref{sec:modeling}, we discussed delexicalization, i.e., not modeling lexical surface forms of pronouns, as one way to counter exclusion in statistical modeling and bucketing. However, a possible counter-argument against this approach is that omitting the surface forms will lead to poor model performance on pronoun-related tasks. We experimentally quantify the loss from (fairer) delexicalization compared to statistical modeling in co-reference resolution.

\subsection{Experimental Setup}
\setlength{\tabcolsep}{10pt}
\begin{table}[!t]\centering
\small
\begin{tabular}{lrrrr}\toprule
& \textbf{Train} & \textbf{Dev} & \textbf{Test} & Total \\
\midrule
\texttt{PRP} & 64,476& 7,881 & 8,067 & 80,424\\
\texttt{PRP\$} & 14,535 & 1,783 & 1,935 & 18,253\\
\midrule
Total & 79,011 & 9,664 & 10,002 & 98,677 \\
\bottomrule
\end{tabular}
\caption{Number of pronoun replacements in the training, development, and test portion of OntoNotes 5.0 for \texttt{PRP} and \texttt{PRP\$}, respectively.}\label{tab:replacements}
\end{table}
\setlength{\tabcolsep}{5.9pt}
\begin{table*}[!t]\centering
\small
\begin{tabular}{lccccccccccccc}\toprule
&\multicolumn{3}{c}{\textbf{MUC}} & \multicolumn{3}{c}{\textbf{CEAF$_{\phi 4}$}} & \multicolumn{3}{c}{\textbf{\textbf{B$^3$}}}  & \multicolumn{3}{c}{\textbf{\textbf{AVG}}}  \\
& \textbf{P} & \textbf{R} & \textbf{F1} &\textbf{P} & \textbf{R} & \textbf{F1} &\textbf{P} & \textbf{R} & \textbf{F1} &\textbf{P} & \textbf{R} & \textbf{F1} \\\cmidrule(lr){2-4}\cmidrule(lr){5-7} \cmidrule(lr){8-10}\cmidrule(lr){11-13}
\citep{dobrovolskii-2021-word} &84.9 &87.9 &86.3 &76.1 &77.1 &76.6 &77.4 &82.6 &79.9 &-- &-- &81.0 \\\cmidrule(lr){2-13}
- reproduction &84.7 &87.5 &86.1 &75.6 &76.7 &76.1 &77.2 &82.0 &79.5 &79.2 &82.1 &80.6 \\
- replace test set &69.7 &70.7 &70.2 &63.2 &49.1 &55.2 &50.1 &56.1 &52.9 &61.0 &58.6 &59.4 \\
\rowcolor{Gray}$\Delta_{\mathit{repl. test-repr.}}$ &-15.0 &-16.8 &-15.9 & -12.4 &-27.6 &-20.9 &-27.1 &-25.9 &-26.6 &-18.2 &-23.5 &-21.2 \\
- replace all &81.6 &83.1 &82.4 &73.08 &72.9 &73.0 &72.3 &75.3 &73.7 &75.7 &77.1 &76.4 \\
\rowcolor{Gray}$\Delta_{\mathit{repl. all-repr.}}$ &-3.1 &-4.4 &-3.7 &-2.5 &-3.8 &-3.1 &-4.9 &-6.7 &-5.8 &-3.5 &-5.0 &-4.2 \\
\bottomrule
\end{tabular}
\caption{Results of the delexicalization experiment. We report the results of the RoBERTa large-based word-level co-reference resolution model as reported by \citet{dobrovolskii-2021-word}, our reproduction, as well as variants trained and/ or  tested on versions of the data set in which we replace the pronouns. All scores were produced using the offical CoNLL-2012 scorer. We report precision (P), recall (R), and F1-score (F1) for MUC, CEAF$_{\phi 4}$, and B$^3$, respectively, as well as the averages (AVG).  The rows highlighted in gray indicate the obtained losses.}\label{tab:results}
\end{table*}
\paragraph{Task, Dataset, and Measures.} 
We resort to co-reference resolution, a task where knowledge about pronouns and related gender identity assumptions play an important role. We use the English portion of the \emph{OntoNotes 5.0} dataset~\citep{weischedel2012ontonotes}, which consists of texts annotated with co-reference information across five domains (news, conversational telephone speech, weblogs, usenet newsgroups, broadcast, and talk shows). We prepare three variants: (i) the first version consists of the plain original data; (ii) in the second variant, we \emph{replace all pronouns} in the test set with the respective part-of-speech token, according to the Penn Treebank Project~\citep{santorini1990part}, i.e., \texttt{PRP} for personal pronouns, and \texttt{PRP\$} for possessive pronouns. Finally, we provide a version (iii) in which we replace pronouns in the train, dev, and test splits. Note that our strategy is pessimistic, as we also replace non-3rd person pronouns, i.e., \emph{I}, \emph{you}, \emph{ourselves}, etc. We show the number of replacements in Table~\ref{tab:replacements}. For scoring, we use the official CoNLL-2012 scorer~\citep{pradhan2012conll}. We report the results in terms of MUC~\citep{vilain1995model}, B$^3$~\citep{bagga1998algorithms}, and CEAF$_{\phi 4}$~\citep{luo-2005-coreference} precision, recall, and F1-measure, as well as the averages across these scores. %

\paragraph{Models and Baselines.} We want to obtain an intuition about the tradeoffs in the delexicalization setup, not to outperform previous results. For this reason, we resort to the recently proposed word-level co-reference model~\citep{dobrovolskii-2021-word}, a highly efficient model competitive with the state-of-the-art. The model consists of a separate co-reference resolution module and a separate span extraction module. In an initial step, we compute token representations from a Transformer~\citep{Transformer}-based encoder through aggregation of initial representations via learnable weights. In a next step, we compute co-reference relationships. To this end, the token representations are passed into an antecedent pruning procedure based on a bilinear scoring function for obtaining $k$ antecedent candidates for each token through coarse-grained scoring. Then an additional feed-forward neural network computes finer-grained scores. The final antecedent score is the sum of these two scores. We select the candidate with the highest score as the antecedent. Negative scores indicate no antecent for a token. Tokens assumed to be part of a co-reference relationship are passed into the span extraction module. The module consists of an additional feed-forward network, which is followed by convolutions with two output channels (for start and end scores). For further details see the original work.
Our baseline is the model trained and evaluated on the original OntoNotes portions (\emph{reproduction}). We compare with the evaluation of this model on the pronoun-replaced test set (\emph{replace test set}) and a version of this model trained on the replaced training set and evaluated on the replaced test set, respectively (\emph{replace all}).

\paragraph{Model Configuration, Training, and Optimization.} We choose RoBERTa large~\citep{liu2019roberta}\footnote{\url{https://huggingface.co/roberta-large}} as the base encoder and fix all other hyperparameters to the ones provided in the original implementation of  \citet{dobrovolskii-2021-word}: the window size is set to $512$ tokens, dropout rate to $0.3$, the learning rate of the encoder is set to $1\cdot 10^{-5}$ and of the task-specific layers to $3\cdot 10^{-4}$, respectively.  %
We train the co-reference module with a combination of the negative log marginal likelihood and binary cross-entropy as an additional regularization factor (weight set to $0.5$). The span extraction module is trained using cross-entropy loss. We optimize the sum of the two losses jointly with Adam~\citep{AdamW} for 20 epochs and apply early stopping based on validation set performance (word-level F1) with a patience of 3 epochs.

\subsection{Results and Discussion}

We show the results in Table~\ref{tab:results}. We are roughly able to reproduce the results reported by \citep{dobrovolskii-2021-word}, confirming the effectiveness of their approach and the validity of our experimental setup. When we replace pronouns in the test set, the results drop massively, with up to $-27.6$ percentage points CEAF$_{\phi 4}$ recall. On average, the results drop by $21.2$ percentage points in F1-measure. This decrease demonstrates the heavy reliance of this model on the lexical surface forms of the pronoun sets seen in the training. However, when we replace the pronouns in the training portion of OntoNotes with the special tokens, we can mitigate these losses by a large margin (losses up to $-5.8$ B$^3$ F1, and on average $-4.2$ F1).
These results are highly encouraging, given that a) we replaced \emph{all} pronouns, including non-third person pronouns, and b) the model has not been trained on these placeholders in the pretraining phase. The model can not rely on possibly discriminating correlations between names or occupations and pronoun sets and will represent neopronouns the same way as it will represent established pronoun sets. So a delexicalization approach can increase fairness in co-reference resolution and retain high system performance, as we can expect even smaller drops from a more careful selection of replacements. %

\section{Conclusion}

This work provides an initial overview of the plethora of current phenomena in 3rd person pronoun usage in the English language. For practical and ethical reasons, the NLP community should acknowledge the broad spectrum of possible identities and the respective manifestations in written and oral communication. Especially since many emerging phenomena are still under-researched, and even while it remains to be seen if and how these become more established ways of referring to individuals. 
Language is consistently evolving, and NLP researchers and practitioners should account for this to provide genuinely inclusive systems. Notably, pronouns, traditionally handled as a close class of words, currently seem to function closer to an open class. 
Based on our observations, which originate from literature research, research in non-academic publicly available writing, as well as a corpus study, we have defined a series of five desiderata and applied those to the discussion of existing and novel modeling paradigms. In this context, we raised the questions \emph{when} and \emph{how} to model pronouns and whether and how to \emph{include users} in these decisions. We consider this document an initial and living draft and hope to start a broader discussion on the topic. Our study can inform future NLP research and beyond and serve as a starting point for creating novel modeling procedures. In the future, we will look at pronoun-related issues within concrete tasks and in multilingual  scenarios.

\section*{Acknowledgments}
The work of Anne Lauscher and Dirk Hovy is funded by the  European Research Council (ERC) under the European Union’s Horizon 2020 research and innovation program (grant agreement No. 949944, INTEGRATOR). We thank Emily Bender and Chandler May for sharing their ideas related to our project. 

\section*{Further Ethical Discussion}
We have described and experimented with phenomena related to third-person pronouns focusing on the English language only. Naturally, this work comes with several limitations. For instance, while we pointed the reader to the variety of pronoun-related phenomena in other languages, a thorough \emph{multilingual and cross-lingual discussion} would have exceeded the scope of this manuscript. This lacuna includes the discussion of neopronouns in other languages.
Similarly, while we acknowledged identities beyond the binary gender as well as otherkin identities, due to our focus on pronouns, we did not investigate \emph{other identity-related terms}. This aspect includes their handling in NLP and the range of issues related to identity-exclusivity. Finally, at the current state of the manuscript, the desiderata discussed are, as reported, based on our expert knowledge, our activities within the LGBTQIA+ community, and informal exchanges with individuals using gender-neutral pronouns. In the future, we will validate these assumptions through a structured survey to present a more inclusive perspective on the discussed issues.

\bibliography{custom}

\begin{thebibliography}{51}
\expandafter\ifx\csname natexlab\endcsname\relax\def\natexlab#1{#1}\fi

\bibitem[{Bagga and Baldwin(1998)}]{bagga1998algorithms}
Amit Bagga and Breck Baldwin. 1998.
\newblock \href
  {https://citeseerx.ist.psu.edu/viewdoc/download?doi=10.1.1.47.5848&rep=rep1&type=pdf}
  {Algorithms for scoring coreference chains}.
\newblock In \emph{Proc. Linguistic Coreference Workshop at the first Conf. on
  Language Resources and Evaluation (LREC)}, pages 563--566, Granada, Spain.

\bibitem[{Barikeri et~al.(2021)Barikeri, Lauscher, Vuli{\'c}, and
  Glava{\v{s}}}]{barikeri-etal-2021-redditbias}
Soumya Barikeri, Anne Lauscher, Ivan Vuli{\'c}, and Goran Glava{\v{s}}. 2021.
\newblock \href {https://doi.org/10.18653/v1/2021.acl-long.151}
  {{R}eddit{B}ias: A real-world resource for bias evaluation and debiasing of
  conversational language models}.
\newblock In \emph{Proceedings of the 59th Annual Meeting of the Association
  for Computational Linguistics and the 11th International Joint Conference on
  Natural Language Processing (Volume 1: Long Papers)}, pages 1941--1955,
  Online. Association for Computational Linguistics.

\bibitem[{Barocas et~al.(2017)Barocas, Crawford, Shapiro, and
  Wallach}]{barocas2017problem}
Solon Barocas, Kate Crawford, Aaron Shapiro, and Hanna Wallach. 2017.
\newblock The problem with bias: Allocative versus representational harms in
  machine learning.
\newblock In \emph{9th Annual Conference of the Special Interest Group for
  Computing, Information and Society}.

\bibitem[{Blodgett et~al.(2020)Blodgett, Barocas, Daum{\'e}~III, and
  Wallach}]{blodgett-etal-2020-language}
Su~Lin Blodgett, Solon Barocas, Hal Daum{\'e}~III, and Hanna Wallach. 2020.
\newblock \href {https://doi.org/10.18653/v1/2020.acl-main.485} {Language
  (technology) is power: A critical survey of {``}bias{''} in {NLP}}.
\newblock In \emph{Proceedings of the 58th Annual Meeting of the Association
  for Computational Linguistics}, pages 5454--5476, Online. Association for
  Computational Linguistics.

\bibitem[{Bolukbasi et~al.(2016)Bolukbasi, Chang, Zou, Saligrama, and
  Kalai}]{Bolukbasi}
Tolga Bolukbasi, Kai{-}Wei Chang, James~Y. Zou, Venkatesh Saligrama, and
  Adam~Tauman Kalai. 2016.
\newblock \href
  {https://proceedings.neurips.cc/paper/2016/hash/a486cd07e4ac3d270571622f4f316ec5-Abstract.html}
  {Man is to computer programmer as woman is to homemaker? debiasing word
  embeddings}.
\newblock In \emph{Advances in Neural Information Processing Systems 29: Annual
  Conference on Neural Information Processing Systems 2016, December 5-10,
  2016, Barcelona, Spain}, pages 4349--4357.

\bibitem[{Bordia and Bowman(2019)}]{bordia-bowman-2019-identifying}
Shikha Bordia and Samuel~R. Bowman. 2019.
\newblock \href {https://doi.org/10.18653/v1/N19-3002} {Identifying and
  reducing gender bias in word-level language models}.
\newblock In \emph{Proceedings of the 2019 Conference of the North {A}merican
  Chapter of the Association for Computational Linguistics: Student Research
  Workshop}, pages 7--15, Minneapolis, Minnesota. Association for Computational
  Linguistics.

\bibitem[{Butler(1990)}]{butler2002gender}
Judith Butler. 1990.
\newblock \emph{Gender trouble}, 1st edition.
\newblock Routledge Classics, New York, NY, USA.

\bibitem[{Butler(2004)}]{butler2004undoing}
Judith Butler. 2004.
\newblock \emph{Undoing Gender}, 1st edition.
\newblock Routledge, New York, NY, USA.

\bibitem[{Cao and Daum{\'e}~III(2021)}]{cao-daume-iii-2021-toward}
Yang~Trista Cao and Hal Daum{\'e}~III. 2021.
\newblock \href {https://doi.org/10.1162/coli_a_00413} {Toward gender-inclusive
  coreference resolution: An analysis of gender and bias throughout the machine
  learning lifecycle*}.
\newblock \emph{Computational Linguistics}, 47(3):615--661.

\bibitem[{Cherry-Reid(2020)}]{cherry2020music}
Katharine~A Cherry-Reid. 2020.
\newblock \href {https://hdl.handle.net/1807/103774} {\emph{Music to Our Ears:
  Using a Queer Folk Song Pedagogy to do Gender and Sexuality Education}}.
\newblock Ph.D. thesis, University of Toronto (Canada).

\bibitem[{Conrod(2019)}]{conrod2019they}
Kirby Conrod. 2019.
\newblock \href
  {https://linguistics.washington.edu/research/graduate/pronouns-raising-and-emerging}
  {\emph{Pronouns Raising and Emerging}}.
\newblock Ph.D. thesis, University of Washington.

\bibitem[{Dev et~al.(2020)Dev, Li, Phillips, and Srikumar}]{dev2020measuring}
Sunipa Dev, Tao Li, Jeff~M. Phillips, and Vivek Srikumar. 2020.
\newblock \href {https://aaai.org/ojs/index.php/AAAI/article/view/6267} {On
  measuring and mitigating biased inferences of word embeddings}.
\newblock In \emph{The Thirty-Fourth {AAAI} Conference on Artificial
  Intelligence, {AAAI} 2020, The Thirty-Second Innovative Applications of
  Artificial Intelligence Conference, {IAAI} 2020, The Tenth {AAAI} Symposium
  on Educational Advances in Artificial Intelligence, {EAAI} 2020, New York,
  NY, USA, February 7-12, 2020}, pages 7659--7666. {AAAI} Press.

\bibitem[{Dev et~al.(2021)Dev, Monajatipoor, Ovalle, Subramonian, Phillips, and
  Chang}]{dev-etal-2021-harms}
Sunipa Dev, Masoud Monajatipoor, Anaelia Ovalle, Arjun Subramonian, Jeff
  Phillips, and Kai-Wei Chang. 2021.
\newblock \href {https://aclanthology.org/2021.emnlp-main.150} {Harms of gender
  exclusivity and challenges in non-binary representation in language
  technologies}.
\newblock In \emph{Proceedings of the 2021 Conference on Empirical Methods in
  Natural Language Processing}, pages 1968--1994, Online and Punta Cana,
  Dominican Republic. Association for Computational Linguistics.

\bibitem[{Devlin et~al.(2019)Devlin, Chang, Lee, and
  Toutanova}]{devlin-etal-2019-bert}
Jacob Devlin, Ming-Wei Chang, Kenton Lee, and Kristina Toutanova. 2019.
\newblock \href {https://doi.org/10.18653/v1/N19-1423} {{BERT}: Pre-training of
  deep bidirectional transformers for language understanding}.
\newblock In \emph{Proceedings of the 2019 Conference of the North {A}merican
  Chapter of the Association for Computational Linguistics: Human Language
  Technologies, Volume 1 (Long and Short Papers)}, pages 4171--4186,
  Minneapolis, Minnesota. Association for Computational Linguistics.

\bibitem[{Dobrovolskii(2021)}]{dobrovolskii-2021-word}
Vladimir Dobrovolskii. 2021.
\newblock \href {https://doi.org/10.18653/v1/2021.emnlp-main.605} {Word-level
  coreference resolution}.
\newblock In \emph{Proceedings of the 2021 Conference on Empirical Methods in
  Natural Language Processing}, pages 7670--7675, Online and Punta Cana,
  Dominican Republic. Association for Computational Linguistics.

\bibitem[{Gonen and Goldberg(2019)}]{gonen-goldberg-2019-lipstick}
Hila Gonen and Yoav Goldberg. 2019.
\newblock \href {https://doi.org/10.18653/v1/N19-1061} {Lipstick on a pig:
  {D}ebiasing methods cover up systematic gender biases in word embeddings but
  do not remove them}.
\newblock In \emph{Proceedings of the 2019 Conference of the North {A}merican
  Chapter of the Association for Computational Linguistics: Human Language
  Technologies, Volume 1 (Long and Short Papers)}, pages 609--614, Minneapolis,
  Minnesota. Association for Computational Linguistics.

\bibitem[{Grandstrand(1998)}]{donath2002identity}
Ove Grandstrand. 1998.
\newblock \href
  {https://www.taylorfrancis.com/chapters/edit/10.4324/9780203194959-11/identity-deception-virtual-community-judith-donath?context=ubx&refId=61b64d57-148b-4c1e-a010-f83c01b34d7f}
  {Identity and deception in the virtual community}.
\newblock In Peter Kollock and Marc Smith, editors, \emph{Communities in
  Cyberspace}, 1st edition, chapter~2. Routledge, London, UK.

\bibitem[{Hercus(1994)}]{hercus1994grammar}
Luise~A Hercus. 1994.
\newblock \emph{A grammar of the Arabana-Wangkangurru language, Lake Eyre
  Basin, South Australia (Pacific linguistics. Series C)}, 1st edition.
\newblock Dept. of Linguistics, Research School of Pacific and Asian Studies,
  Australian National University.

\bibitem[{Hobbs(1978)}]{hobbs1978resolving}
Jerry~R Hobbs. 1978.
\newblock \href
  {https://www.sciencedirect.com/science/article/pii/0024384178900062}
  {Resolving pronoun references}.
\newblock \emph{Lingua}, 44(4):311--338.

\bibitem[{Hung et~al.(2021)Hung, Lauscher, Ponzetto, and
  Glava{\v{s}}}]{hung2021ds}
Chia-Chien Hung, Anne Lauscher, Simone~Paolo Ponzetto, and Goran Glava{\v{s}}.
  2021.
\newblock \href {https://arxiv.org/abs/2110.08395?context=cs} {{DS-TOD}:
  {E}fficient domain specialization for task oriented dialog}.
\newblock \emph{arXiv preprint arXiv:2110.08395}.

\bibitem[{Kashima and Kashima(1998)}]{doi:10.1177/0022022198293005}
Emiko~S. Kashima and Yoshihisa Kashima. 1998.
\newblock \href {https://doi.org/10.1177/0022022198293005} {Culture and
  language: The case of cultural dimensionsand personal pronoun use}.
\newblock \emph{Journal of Cross-Cultural Psychology}, 29(3):461--486.

\bibitem[{Kingma and Ba(2015)}]{AdamW}
Diederik~P. Kingma and Jimmy Ba. 2015.
\newblock \href {http://arxiv.org/abs/1412.6980} {Adam: {A} method for
  stochastic optimization}.
\newblock In \emph{3rd International Conference on Learning Representations,
  {ICLR} 2015, San Diego, CA, USA, May 7-9, 2015, Conference Track
  Proceedings}.

\bibitem[{Konnelly and Cowper(2020)}]{Konnelly-Cowper-singularthey}
Lex Konnelly and Elizabeth Cowper. 2020.
\newblock \href {https://www.glossa-journal.org/article/id/5288/} {Gender
  diversity and morphosyntax: An account of singular they}.
\newblock \emph{Glossa: a journal of general linguistics}, 5(1).

\bibitem[{Krauthamer(2021)}]{krauthamer2020great}
Helene~Seltzer Krauthamer. 2021.
\newblock \emph{The Great Pronoun Shift: The Big Impact of Little Parts of
  Speech}, 1st edition.
\newblock Routledge.

\bibitem[{Kurita et~al.(2019)Kurita, Vyas, Pareek, Black, and
  Tsvetkov}]{kurita-etal-2019-measuring}
Keita Kurita, Nidhi Vyas, Ayush Pareek, Alan~W Black, and Yulia Tsvetkov. 2019.
\newblock \href {https://doi.org/10.18653/v1/W19-3823} {Measuring bias in
  contextualized word representations}.
\newblock In \emph{Proceedings of the First Workshop on Gender Bias in Natural
  Language Processing}, pages 166--172, Florence, Italy. Association for
  Computational Linguistics.

\bibitem[{Lauscher et~al.(2020)Lauscher, Glava{\v{s}}, Ponzetto, and
  Vuli{\'c}}]{DEBIE}
Anne Lauscher, Goran Glava{\v{s}}, Simone~Paolo Ponzetto, and Ivan Vuli{\'c}.
  2020.
\newblock \href {https://ojs.aaai.org//index.php/AAAI/article/view/6325} {A
  general framework for implicit and explicit debiasing of distributional word
  vector spaces}.
\newblock In \emph{Proceedings of the AAAI Conference on Artificial
  Intelligence}, pages 8131--8138.

\bibitem[{Lauscher et~al.(2021)Lauscher, Lueken, and
  Glava{\v{s}}}]{lauscher2021sustainable}
Anne Lauscher, Tobias Lueken, and Goran Glava{\v{s}}. 2021.
\newblock \href {https://aclanthology.org/2021.findings-emnlp.411} {Sustainable
  modular debiasing of language models}.
\newblock In \emph{Findings of the Association for Computational Linguistics:
  EMNLP 2021}, pages 4782--4797, Punta Cana, Dominican Republic. Association
  for Computational Linguistics.

\bibitem[{Laycock(2012)}]{laycock2012we}
Joseph~P Laycock. 2012.
\newblock \href
  {https://online.ucpress.edu/nr/article-abstract/15/3/65/70585/We-Are-Spirits-of-Another-Sort-Ontological?redirectedFrom=fulltext}
  {“{W}e are spirits of another sort”: Ontological rebellion and religious
  dimensions of the otherkin community}.
\newblock \emph{Nova Religio: The Journal of Alternative and Emergent
  Religions}, 15(3):65--90.

\bibitem[{Liu et~al.(2019)Liu, Ott, Goyal, Du, Joshi, Chen, Levy, Lewis,
  Zettlemoyer, and Stoyanov}]{liu2019roberta}
Yinhan Liu, Myle Ott, Naman Goyal, Jingfei Du, Mandar Joshi, Danqi Chen, Omer
  Levy, Mike Lewis, Luke Zettlemoyer, and Veselin Stoyanov. 2019.
\newblock \href {https://arxiv.org/pdf/1907.11692.pdf} {Roberta: A robustly
  optimized bert pretraining approach}.
\newblock \emph{arXiv preprint arXiv:1907.11692}.

\bibitem[{Luo(2005)}]{luo-2005-coreference}
Xiaoqiang Luo. 2005.
\newblock \href {https://aclanthology.org/H05-1004} {On coreference resolution
  performance metrics}.
\newblock In \emph{Proceedings of Human Language Technology Conference and
  Conference on Empirical Methods in Natural Language Processing}, pages
  25--32, Vancouver, British Columbia, Canada. Association for Computational
  Linguistics.

\bibitem[{Maalouf(2000)}]{maalouf2011identity}
Amin Maalouf. 2000.
\newblock \emph{On identity}, 1st, translated by barbara bray edition.
\newblock Vintage.

\bibitem[{McDonald et~al.(2011)McDonald, Petrov, and
  Hall}]{mcdonald-etal-2011-multi}
Ryan McDonald, Slav Petrov, and Keith Hall. 2011.
\newblock \href {https://aclanthology.org/D11-1006} {Multi-source transfer of
  delexicalized dependency parsers}.
\newblock In \emph{Proceedings of the 2011 Conference on Empirical Methods in
  Natural Language Processing}, pages 62--72, Edinburgh, Scotland, UK.
  Association for Computational Linguistics.

\bibitem[{McGaughey(2020)}]{mcgaughey2020understanding}
Sebastian McGaughey. 2020.
\newblock \href {https://glreview.org/article/understanding-neopronouns/}
  {Understanding neopronouns}.
\newblock \emph{The Gay \& Lesbian Review Worldwide}, 27(2):27--29.

\bibitem[{McKay(1993)}]{mckay1993term}
John~C McKay. 1993.
\newblock \href {https://www.jstor.org/stable/479880} {On the term ``pronoun''
  in italian grammars}.
\newblock \emph{Italica}, 70(2):168--181.

\bibitem[{Miltersen(2016)}]{miltersen2016nounself}
Ehm~Hjorth Miltersen. 2016.
\newblock \href {https://tidsskrift.dk/lwo/article/view/23431} {Nounself
  pronouns: 3rd person personal pronouns as identity expression}.
\newblock \emph{Journal of Language Works-Sprogvidenskabeligt
  Studentertidsskrift}, 1(1):37--62.

\bibitem[{Muqit(2012)}]{muqit2012ideology}
Abd Muqit. 2012.
\newblock \href {http://www.ijssh.org/papers/171-A10046.pdf} {Ideology and
  power relation reflected in the use of pronoun in osama bin laden's speech
  text}.
\newblock \emph{International Journal of Social Science and Humanity},
  2(6):557.

\bibitem[{Pennington et~al.(2014)Pennington, Socher, and
  Manning}]{pennington2014glove}
Jeffrey Pennington, Richard Socher, and Christopher Manning. 2014.
\newblock \href {https://doi.org/10.3115/v1/D14-1162} {{G}lo{V}e: Global
  vectors for word representation}.
\newblock In \emph{Proceedings of the 2014 Conference on Empirical Methods in
  Natural Language Processing ({EMNLP})}, pages 1532--1543, Doha, Qatar.
  Association for Computational Linguistics.

\bibitem[{Postal et~al.(1969)Postal, Reibel, and Schane}]{postal1969so}
Paul Postal, David~A Reibel, and Sanford~A Schane. 1969.
\newblock On so-called pronouns in english.
\newblock \emph{Readings in English transformational grammar}, pages 12--25.

\bibitem[{Pradhan et~al.(2012)Pradhan, Moschitti, Xue, Uryupina, and
  Zhang}]{pradhan2012conll}
Sameer Pradhan, Alessandro Moschitti, Nianwen Xue, Olga Uryupina, and Yuchen
  Zhang. 2012.
\newblock \href {https://aclanthology.org/W12-4501.pdf} {Conll-2012 shared
  task: Modeling multilingual unrestricted coreference in ontonotes}.
\newblock In \emph{Joint Conference on EMNLP and CoNLL-Shared Task}, pages
  1--40.

\bibitem[{Raymond(2016)}]{raymond2016linguistic}
Chase~Wesley Raymond. 2016.
\newblock \href {https://muse.jhu.edu/article/629765} {Linguistic reference in
  the negotiation of identity and action: Revisiting the t/v distinction}.
\newblock \emph{Language}, 92:636--670.

\bibitem[{Roberts(2020)}]{wordofyear}
Julie Roberts. 2020.
\newblock \href
  {https://www.americandialect.org/wp-content/uploads/2019-Word-of-the-Year-PRESS-RELEASE.pdf}
  {2019 word of the year is “(my) pronouns,” word of the decade is singular
  “they” as voted by american dialect society}.
\newblock Press Release, American Dialect Society.

\bibitem[{Rudinger et~al.(2018)Rudinger, Naradowsky, Leonard, and
  Van~Durme}]{rudinger-etal-2018-gender}
Rachel Rudinger, Jason Naradowsky, Brian Leonard, and Benjamin Van~Durme. 2018.
\newblock \href {https://doi.org/10.18653/v1/N18-2002} {Gender bias in
  coreference resolution}.
\newblock In \emph{Proceedings of the 2018 Conference of the North {A}merican
  Chapter of the Association for Computational Linguistics: Human Language
  Technologies, Volume 2 (Short Papers)}, pages 8--14, New Orleans, Louisiana.
  Association for Computational Linguistics.

\bibitem[{Santorini(1990)}]{santorini1990part}
Beatrice Santorini. 1990.
\newblock \href
  {https://repository.upenn.edu/cgi/viewcontent.cgi?article=1603&context=cis_reports}
  {Part-of-speech tagging guidelines for the penn treebank project}.

\bibitem[{Shah et~al.(2020)Shah, Schwartz, and
  Hovy}]{shah-etal-2020-predictive}
Deven~Santosh Shah, H.~Andrew Schwartz, and Dirk Hovy. 2020.
\newblock \href {https://doi.org/10.18653/v1/2020.acl-main.468} {Predictive
  biases in natural language processing models: A conceptual framework and
  overview}.
\newblock In \emph{Proceedings of the 58th Annual Meeting of the Association
  for Computational Linguistics}, pages 5248--5264, Online. Association for
  Computational Linguistics.

\bibitem[{Spivak(1990)}]{spivak1990joy}
Michael Spivak. 1990.
\newblock \emph{The Joy of TeX: A Gourmet Guide to Typesetting with the AMSTeX
  Macro Package}, 2nd edition.
\newblock American Mathematical Society.

\bibitem[{Stryker(2017)}]{stryker2017transgender}
Susan Stryker. 2017.
\newblock \emph{Transgender history: The roots of today's revolution}, 2nd
  edition.
\newblock Seal Press.

\bibitem[{Suntwal et~al.(2019)Suntwal, Paul, Sharp, and
  Surdeanu}]{suntwal-etal-2019-importance}
Sandeep Suntwal, Mithun Paul, Rebecca Sharp, and Mihai Surdeanu. 2019.
\newblock \href {https://doi.org/10.18653/v1/D19-1340} {On the importance of
  delexicalization for fact verification}.
\newblock In \emph{Proceedings of the 2019 Conference on Empirical Methods in
  Natural Language Processing and the 9th International Joint Conference on
  Natural Language Processing (EMNLP-IJCNLP)}, pages 3413--3418, Hong Kong,
  China. Association for Computational Linguistics.

\bibitem[{Vaswani et~al.(2017)Vaswani, Shazeer, Parmar, Uszkoreit, Jones,
  Gomez, Kaiser, and Polosukhin}]{Transformer}
Ashish Vaswani, Noam Shazeer, Niki Parmar, Jakob Uszkoreit, Llion Jones,
  Aidan~N. Gomez, Lukasz Kaiser, and Illia Polosukhin. 2017.
\newblock \href
  {https://proceedings.neurips.cc/paper/2017/hash/3f5ee243547dee91fbd053c1c4a845aa-Abstract.html}
  {Attention is all you need}.
\newblock In \emph{Advances in Neural Information Processing Systems 30: Annual
  Conference on Neural Information Processing Systems 2017, December 4-9, 2017,
  Long Beach, CA, {USA}}, pages 5998--6008.

\bibitem[{Vilain et~al.(1995)Vilain, Burger, Aberdeen, Connolly, and
  Hirschman}]{vilain1995model}
Marc Vilain, John~D Burger, John Aberdeen, Dennis Connolly, and Lynette
  Hirschman. 1995.
\newblock \href {https://aclanthology.org/M95-1005.pdf} {A model-theoretic
  coreference scoring scheme}.
\newblock In \emph{Sixth Message Understanding Conference (MUC-6): Proceedings
  of a Conference Held in Columbia, Maryland, November 6-8, 1995}.

\bibitem[{Webster et~al.(2018)Webster, Recasens, Axelrod, and
  Baldridge}]{webster-etal-2018-mind}
Kellie Webster, Marta Recasens, Vera Axelrod, and Jason Baldridge. 2018.
\newblock \href {https://doi.org/10.1162/tacl_a_00240} {Mind the {GAP}: A
  balanced corpus of gendered ambiguous pronouns}.
\newblock \emph{Transactions of the Association for Computational Linguistics},
  6:605--617.

\bibitem[{Weischedel et~al.(2012)Weischedel, Pradhan, Ramshaw, Kaufman,
  Franchini, El-Bachouti, Xue, Palmer, Hwang, Bonial
  et~al.}]{weischedel2012ontonotes}
Ralph Weischedel, Sameer Pradhan, Lance Ramshaw, Jeff Kaufman, Michelle
  Franchini, Mohammed El-Bachouti, Nianwen Xue, Martha Palmer, Jena~D Hwang,
  Claire Bonial, et~al. 2012.
\newblock \href {https://catalog.ldc.upenn.edu/LDC2013T19} {Ontonotes release
  5.0}.

\end{thebibliography}
\bibliographystyle{acl_natbib}

\end{document}